\pdfoutput=1
\documentclass[11pt]{article}

\usepackage[]{ACL2023}

\usepackage{times} 
\usepackage{latexsym}
\usepackage{listings} 
\usepackage{xcolor}
\usepackage{tcolorbox}
\usepackage{multicol} 
\usepackage{booktabs}
\usepackage{array}
\usepackage{tabularx}

\lstdefinestyle{mystyle}{
    language=Python,
    basicstyle=\ttfamily\small,
    commentstyle=\color{gray},
    keywordstyle=\color{orange},
    stringstyle=\color{blue},
    showstringspaces=false,
    breaklines=true,
    frame=single,
    frameround=tttt, % Add rounded corners
    backgroundcolor=\color{gray!10},
    framexleftmargin=10pt, % Control left margin
    framexrightmargin=10pt % Control right margin
}
\lstdefinelanguage{json}{
    basicstyle=\small\ttfamily,
    columns=fullflexible,
    breaklines=true,
    showstringspaces=false,
    commentstyle=\color{gray},
    keywordstyle=\color{blue},
    stringstyle=\color{black},
    numberstyle=\color{red},
    morekeywords={question, visualization, rationale},
    literate=%
        *{0}{{{\color{red}0}}}{1}
        {1}{{{\color{red}1}}}{1}
        {2}{{{\color{red}2}}}{1}
        {3}{{{\color{red}3}}}{1}
        {4}{{{\color{red}4}}}{1}
        {5}{{{\color{red}5}}}{1}
        {6}{{{\color{red}6}}}{1}
        {7}{{{\color{red}7}}}{1}
        {8}{{{\color{red}8}}}{1}
        {9}{{{\color{red}9}}}{1}
        {:}{{{\color{black}: }}}{1}
        {,}{{{\color{black}, }}}{1},
}

\usepackage[inline]{enumitem}

\usepackage[T1]{fontenc}
\usepackage[utf8]{inputenc}
\usepackage{graphicx} 
\usepackage{amssymb}
\usepackage{mathptmx}  
\usepackage{bookmark}

\graphicspath{{figs/}{figures/}{pictures/}{images/}{./}} % where to search for the images

\usepackage{tabu}                      % only used for the table example
\usepackage{booktabs}                  % only used for the table example
\usepackage{lipsum}                    % used to generate placeholder text
\usepackage{mwe}                       % used to generate placeholder figures

\usepackage{amsmath}

\usepackage{microtype}

\usepackage{inconsolata}

\title{LIDA: A Tool for Automatic Generation of Grammar-Agnostic Visualizations and Infographics using Large Language Models}

\author{Victor Dibia \\
  Microsoft Research \\ 
  \texttt{victordibia@microsoft.com} }

\begin{document}
\maketitle

\newcommand{\lida}{\textsc{lida}} 
\newcommand{\igm}{\textsc{igm}} 
\newcommand{\nl}{\textsc{nl}} 
\newcommand{\llm}{\textsc{llm}} 
\newcommand{\goal}{\textsc{goal explorer}} 
\newcommand{\summarizer}{\textsc{summarizer}}
\newcommand{\hypothesis}{\textsc{Hypothesis}}
\newcommand{\visgen}{\textsc{visGenerator}}
\newcommand{\ui}{\textsc{User Interface}}
\newcommand{\autoviz}{\textsc{autoviz}}
\newcommand{\ver}{\textsc{ver}}
\newcommand{\sevq}{\textsc{sevq}}
\newcommand{\vizops}{\textsc{VizOps}}
\newcommand{\infographics}{\textsc{infographer}}

\begin{abstract}
  Systems that support users in the automatic creation of visualizations must address several subtasks - understand the semantics of data, enumerate relevant visualization goals and generate visualization specifications. In this work, we pose visualization generation as a multi-stage generation problem and argue that well-orchestrated pipelines based on large language models (\llm{}s) and image generation models (\igm{}s) are suitable to addressing these tasks. We present \lida{}, a novel tool for generating grammar-agnostic visua\textbf{\underline{li}}zations an\textbf{\underline{d}} infogr\textbf{\underline{a}}phics. \lida{} comprises of 4 modules - A \summarizer{} that converts data into a rich but compact natural language summary, a \goal{} that enumerates visualization goals given the data, a \visgen{} that generates, refines, executes and filters visualization code and an \infographics{} module that yields data-faithful stylized graphics using \igm{}s. \lida{} provides a python api, and a \textit{hybrid} \ui{} (direct manipulation and \textit{multilingual} natural language) for interactive chart, infographics and data story generation. Code and demo are  available at this url - \url{https://microsoft.github.io/lida/}
\end{abstract}

\section{Introduction}
\label{sec:intro}

\begin{figure*}[ht]
    \centering
    \includegraphics[width=\textwidth]{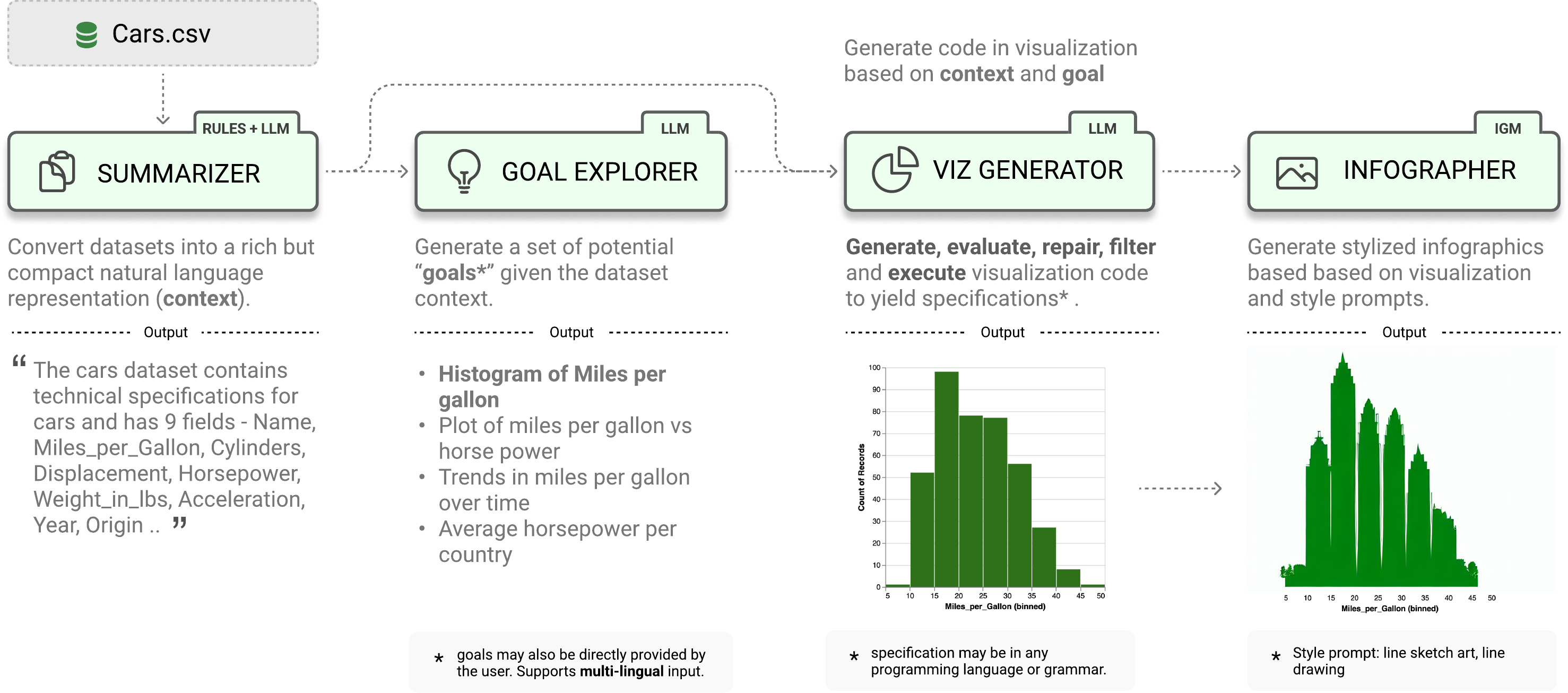}
    \caption{ \lida{} generates visualizations and infographics across 4 modules - data summarization, goal exploration, visualization generation and infographics generations. Example output from each module is shown.}
    \label{fig:systemmodules}
\end{figure*}

\begin{figure*}[ht]
    \centering
    \includegraphics[width=\textwidth]{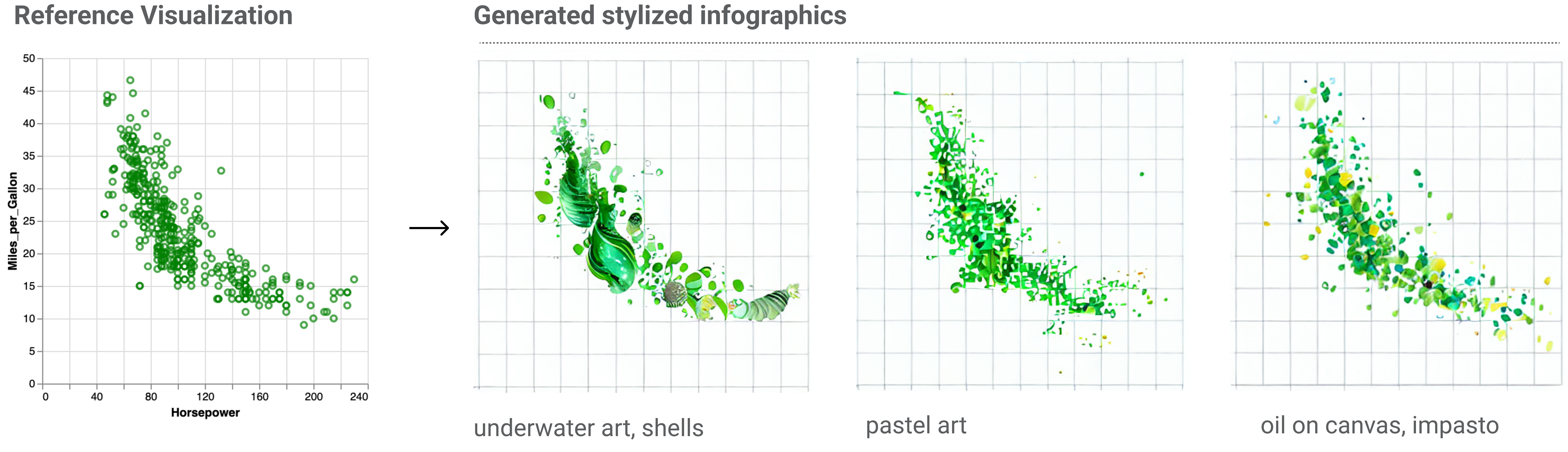}
    \caption{ Example \textit{data-faithful} infographics  and associated style prompts generated with \lida{}. }
    \label{fig:infographics_small}
\end{figure*}

% \begin{figure*}[ht]
%     \centering
%     \includegraphics[width=\textwidth]{figs/lidarch.pdf}
%     \caption{Architecture of \lida{}.}
%     \label{fig:system}
% \end{figure*}

Visualizations make data accessible by reducing the cognitive burden associated with extracting insights from large tabular datasets. However, visualization authoring is a complex creative task, involving multiple steps. First the user must build familiarity with the  dataset (content and semantics) and enumerate a set of relevant goals or hypotheses that can be addressed using the data. Next, users must  select the right visualization representation (marks, transformations and layout) for each goal. Finally, the user must implement the visualization either as code or using available direct manipulation interfaces. Each of these steps require expertise, and can be tedious as well as error prone for \textit{users with limited visualization experience} (novices).  Existing research has sought to address these challenges by \textit{automating} the visualization (\autoviz{}) creation process, given a dataset \cite{podo2023machine}. \textit{Automation} may occur in two modes: i.) fully automated - the system automatically generates visualizations relevant to the data ii.) semi-automated - the user specifies their goals and the system generates visualizations that address these goals. The former mode is valuable for users unfamiliar with the data and the latter is valuable for users with some familiarity with the data and the visualization task.

% review marks transofrmation and layout 

Consequently, a successful \autoviz{} tool must excel at each of several \textit{subtasks} - understand the semantics of the data, enumerate relevant visualization goals and generate visualization specifications that meet  syntax, design, task and perceptual requirements of these goals \cite{podo2023machine}. Furthermore, given the target demographic (novice users), such a tool must support the user by offering \nl{} (\nl{}) interaction modalities \cite{mitra2022facilitating,narechania2020nl4dv,chen2022type}, affordances to control system behavior and sense making tools to understand and debug/verify system behavior. While related work has addressed aspects of the \autoviz{} task, there are several known limitations \cite{podo2023machine} such as they: \begin{enumerate*}[label=(\roman*)]
    \item rely on heuristics that are limited in coverage, challenging to craft and tedious to maintain \cite{Wongsuphasawat_2017}. 
    \item require significant user interaction to generate visualizations \cite{Wongsuphasawat_2017,moritz2018formalizing}.
    \item implement automated approaches that offer limited control over system input and output \cite{dibia2019data2vis}
    \item require grammar (or chart type) specific  training data and model architectures \cite{dibia2019data2vis,Yuyu:2018:DeepEye} for each sub task,
    \item do not consider alternative chart representation formats such as infographics. 
  \end{enumerate*} 
  
 Concurrently, advances in large foundation models \cite{bommasani2021opportunities} have shown state of the art performance on a variety of creative tasks such as multilingual text generation, code generation, image captioning, image generation, and image editing. In this work, we argue that the vast capabilities of these models can be \textit{assembled} to address the \autoviz{} task, whilst \textit{addressing the limitations of existing approaches}. This work makes the following contributions:   

\begin{itemize}[leftmargin=.2in]
    \item We present a novel multi-stage, modular approach (Fig \ref*{fig:systemmodules}) for the automatic generation of data visualization and infographics using \llm{}s\footnote{This work primarily utilizes the OpenAI  \textit{gpt-3.5-turbo-x} line of models for text and code generation.}. Specifically, we \begin{enumerate*}[label=(\roman*)]
        \item Efficiently represent datasets as \nl{} summaries, suitable as grounding context for an \llm{} to address visualization tasks.
        \item Generate a set of visualization goals using \llm{}s. Importantly, we leverage prompt engineering to steer the model towards generating \textit{correct} visualization that follow \textit{best practices} (see Appendix \ref{sec:reflection}).
        \item Apply \llm{}s to generate grammar-agnostic visualization specification based on generated (or human provided) goals. 
        \item Provide a \textit{hybrid interface} that supports traditional direct manipulation controls (e.g., manually select which fields to explore) and a rich \textbf{multilingual} \nl{} interface to support user's with varied skill/experience.
       
        \item Apply text-conditioned image generation models (\igm{}) models in generating stylized infographics that are both informative (generally faithful to data), aesthetically pleasing, memorable and engaging (see section \ref{sec:related:infographics}).
      \end{enumerate*} 
    \item We introduce metrics for evaluating \llm{}-enabled visualization tools, including a metric for pipeline reliability (visualization error rate - \ver{}), and visualization quality (self-evaluated visualization quality - \sevq{}) (see section \ref{sec:evaluation}).
    \item We implement our approach in an Open Source library - \lida{}\footnote{https://microsoft.github.io/lida/.}. \lida{} provides a python api, a web api and a \textbf{rich web interface} useful for research and practical applications.
    % that supports multiple affordances such as summary enrichment, conversational generation and editing of visualizations, infographics and data stories. 

\end{itemize}

\noindent Compared to existing \autoviz{} approaches, \lida{} proposes an implementation that is \textbf{simplified} (eliminates the need for subtask-specific models), \textbf{general} (can be adapted to generate visualizations in any programming language or grammar), \textbf{flexible} (individual modules can be optimized) and \textbf{scalable} (the system performance will \textit{improve} with advances in the underlying \llm{}). Taken together, these contributions provide building blocks towards complex workflows such as \text{visualization translation}, \textit{chart question answering} (with applications in accessibility of charts), automated \textit{data exploration} and \textit{automated data stories}.  

To the best of our knowledge, \lida{} is the first tool to formulate visualization/infographic generation as a multi-step generation task and demonstrate an end-to-end pipeline that addresses a variety of subtasks.

\section{Related Work}
\label{sec:related}

\lida{} is informed by research on large foundation models applied to creative tasks across modalities such as text and images, and advances in automated generation of visualizations and infographics.  

\subsection{Foundation Models for Creative Tasks} 
\label{sec:related:llm}

Advances in large transformer-based \cite{vaswani2017attention} models trained on massive amounts of data (terabytes of text and images) have led to a paradigm shift where a single model demonstrates state of the art task performance across multiple data modalities such as text, images, audio and video. These models, also known as foundation models \cite{bommasani2021opportunities},  have been shown to be effective for a variety of \textit{human creativity} tasks. \llm{}s like the GPT3 series \cite{brown2020language}, OPT \cite{zhang2022opt}, PALM \cite{chowdhery2022palm}, LAMBDA \cite{lambda}   learn complex semantics of language allowing them to be effective in tasks such as text summarization, question answering. Code \llm{}s such as Codex \cite{chen2021evaluatingcodex}, AlphaCode \cite{li2022competitionalphacode}, InCoder \cite{fried2022incoder} show state of the art performance on a suite of code intelligence tasks. Finally, models such as CLIP \cite{radford2021learning}, DALLE \cite{ramesh2022hierarchical, ramesh2021zero} and Latent Diffusion \cite{rombach2022high} have shown state of the art capabilities on image generation tasks such as image captioning, image editing, and image generation. 

In this work, we adopt insights from Program-Aided Language models \cite{gao2022pal} - a setup where \llm{}s generate programs as the intermediate reasoning steps, but offload the solution step to a runtime such as a python interpreter. We leverage the \textit{language modeling} capabilities of \llm{}s in generating \textit{semantically meaningful} visualization goals, and their \textit{code writing} capabilities in generating \textit{visualization code} which is compiled to yield visualizations. These visualizations (images) are then used as input to image generation models in generating stylized infographics.

\subsection{ Automated Visualization (\autoviz{})}
\label{sec:related:visgen}

Extant \autoviz{} research have explored multiple approaches such as heuristics, task decomposition or learning based approaches. Heuristics-based approaches explore properties of data in generating a search space of potential visualizations \cite{Wongsuphasawat_2017}, ranking these visualizations based on quality attributes \cite{Yuyu:2018:DeepEye, moritz2018formalizing} and presenting them to the user. For example, DeepEye \cite{Yuyu:2018:DeepEye} enumerates all  possible visualizations and classifies/ranks them as “good” or “bad” using a binary decision tree classifier while Voyager \cite{Wongsuphasawat_2017} uses heuristics to enumerate the space of visualizations. However, heuristics can be tedious to maintain, may have poor coverage of the visualization space and does not leverage information encoded in existing datasets. More recent work has explored a task decomposition approach where the \autoviz{} process is decomposed into multiple tasks that are solved individually via specialized tools and aggregated to yield visualizations \cite{narechania2020nl4dv,chen2022type, wang2022towards}. For example NL4DV \cite{narechania2020nl4dv} implements a custom query engine that parses natural language queries, identifies attributes/tasks and generates Vega-Lite specifications. A limitation of task decomposition approaches is that they are bottlenecked by the implementation performance for each step (e.g., limitations with models for disambiguating natural language queries as seen in NL4DV \cite{narechania2020nl4dv}). Finally, end-to-end learning-based approaches seek to automatically learn mappings from data directly to generated visualizations. For example, Data2Vis \cite{dibia2019data2vis} (the most relevant work to this study) uses a sequence to sequence model that implicitly addresses \autoviz{} subtasks by learning a mapping from raw JSON data sampled from datasets to Vega-Lite \cite{satyanarayan2017vegalite} specifications. Some limitations of current learning approaches is that they are limited to a single grammar, require custom models, custom paired training data  and training objectives \cite{dibia2019data2vis,Yuyu:2018:DeepEye, chen2022type} for each supported grammar, and do not provide a path to generating infographics. Furthermore, they do not provide mechanisms for fine-grained control of visualization output or provide robust error detection and recovery strategies.

\lida{}  addresses these limitations in several ways: 
\begin{enumerate*}[label=(\roman*)]
    \item Leverages patterns learned by \llm{}s from massive language and code dataset, applying this knowledge to subtasks.
    \item Provides a single grammar-agnostic pipeline that generates visualization in multiple programming languages and visualization grammars.
    \item Supports natural language based control of generated visualizations.  
    \item leverage emergent capabilities of large language models such  chain of thought reasoning to improve reliability of generated text/code \cite{kojima2022large,wei2022chain, shi2022language}, model calibration \cite{kadavath2022language} (predictions on correctness probabilities of visualizations) as well as self-consistency \cite{wang2022self} in ranking/filtering results.
    \item provides a mechanism for generating infographics that are data-faithful and aesthetically pleasing. 
    \item supports a fully automatic mode where an \llm{} is used to discover meaningful goals/hypotheses (fields to visualize, questions to ask) or a semi automatic mode where the user provides a hypothesis and it generates a visualization.
  \end{enumerate*} 
  
  \noindent By choosing to cast visualization/infographic generation as generation tasks that offloads core problem solving to \llm{}s and \igm{}s, \lida{} simplifies the design and maintenance of such systems.

\subsection{Infographics Generation}
\label{sec:related:infographics}

Infographics (information graphics) are visual artifacts that seek to convey complex data-driven narratives using visual imagery and embellishments \cite{harrison2015infographic}.  Existing research has shown that infographics are aesthetically pleasing, engaging and more memorable \cite{tyagi2021user,harrison2015infographic,haroz2015isotype}, at no additional cost to the user \cite{haroz2015isotype}. These properties have driven their applications in domains like fashion, advertisemnt, business and general communications. However, the creation of infographics that convey data insights can be a tedious process for content creators, often requiring skills across multiple tools and domains.  Research on infographic generation have mainly explored the creation of pictographs \cite{haroz2015isotype} - replacing the marks on traditional charts with generated images and learning to extract/transfer styles from existing pictographs \cite{shi2022supporting}. In this work, we extend this domain to exploring the generation of both visual marks as well as generating the entire infographic based on natural language style descriptions using large image generation models such as DALLE \cite{ramesh2022hierarchical, ramesh2021zero} and Latent Diffusion \cite{rombach2022high}. This approach also enables user-generated visual styles and personalization of visualizations to fit user preferences such as color palettes, visual styles, fonts etc.

\section{The \lida{} System}
\label{sec:system}

\lida{} comprises of 4 core modules - a \summarizer{}, a \goal{}, a \visgen{} and an \infographics{} (see Fig \ref*{fig:systemmodules}). Each module is implemented in the \lida{}  \href{https://github.com/microsoft/lida}{github repo} as a python library with an optional user interface (see Appendix \ref{sec:appendix_user_interface}). 
 
\subsection{\summarizer{}}
\label{sec:summarizer}
\begin{figure}[htbp]
    \centering
    \includegraphics[width=\columnwidth]{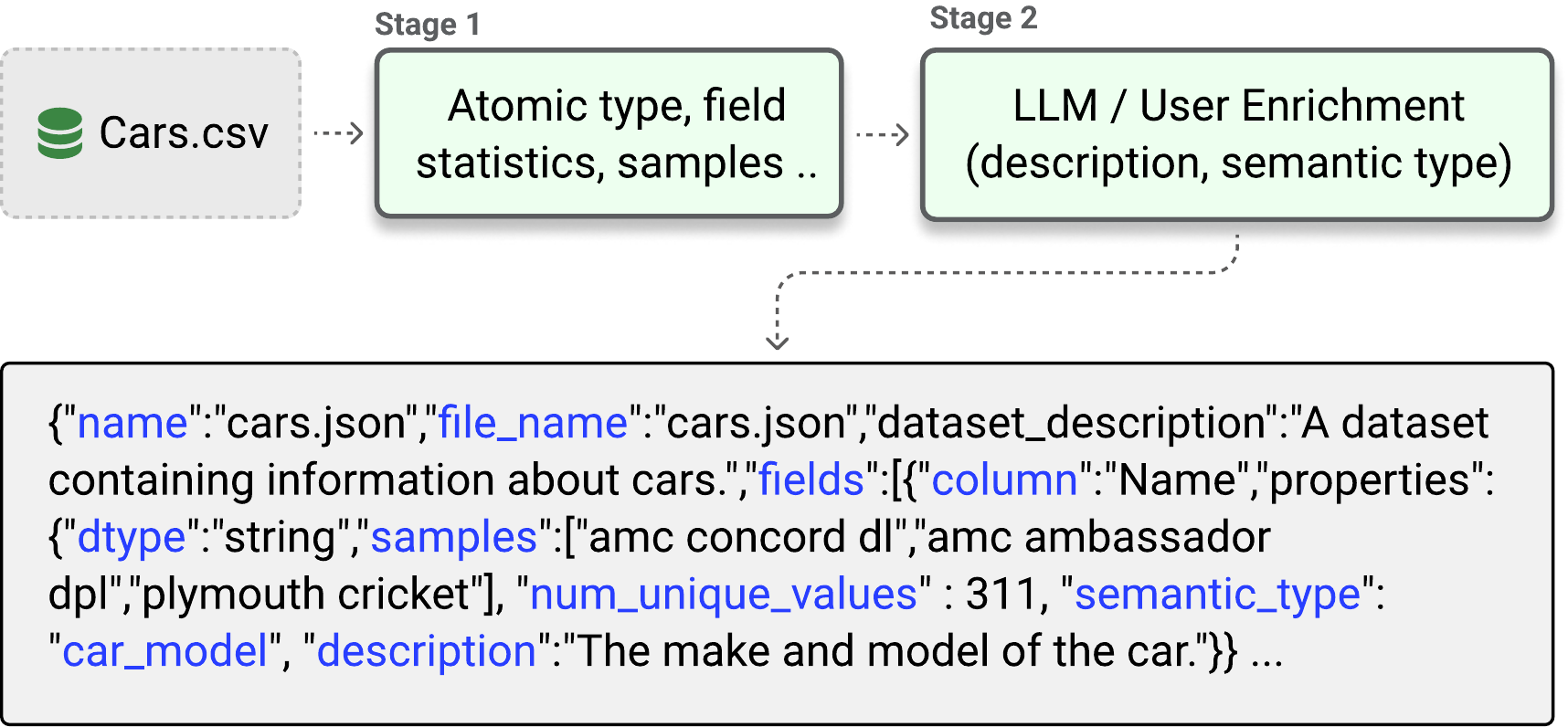}
    \caption{ The \summarizer{} module constructs a \nl{} summary from extracted data properties (atomic types, field statistics) and an optional LLM enrichment (predicted field descriptions, semantic types).}
    \label{fig:summarizer}
\end{figure}

\noindent \llm{}s are capable zero shot predictors, able to solve multiple tasks with little or no guiding examples. However, they can suffer from hallucination e.g., generating text that is not grounded in training data or the current task. One way to address this is to \textit{augment} \cite{mialon2023augmented} the \llm{} with grounding context. Thus, the goal of the summarizer is to produce an \textbf{information dense} but \textbf{compact} \footnote{ Note: the summary must be compact in order to maximize the limited context token budget of \llm{s}.} summary for a given dataset that is \textit{useful as grounding context} for visualization tasks. A useful context is defined as one that \textit{contains information an analyst would need to understand the dataset and the tasks that can be performed on it}. The summary is implemented in two stages (see Fig \ref{fig:summarizer})

\noindent \textbf{Stage 1 - Base summary generation}: We apply rules in extracting dataset properties including atomic types (e.g., integer, string, boolean) using the pandas library \cite{pandas-mckinney-proc-scipy-2010}, general statistics (min, max, \# unique values) and a random non-null list of $n$ samples  for each column.

\noindent \textbf{Stage 2 - Summary enrichment}: The base summary is optionally enriched by an \llm{} or a user via the \hyperref[sec:ui]{\lida{} ui} to include semantic description of the dataset (e.g., a dataset on the technical specification of cars), and fields (e.g., miles per gallon for each car) as well as field semantic type prediction \cite{zhang2019sato}.

\subsection{\goal{} }

\begin{figure}[htbp]
  \centering
  \includegraphics[width=\columnwidth]{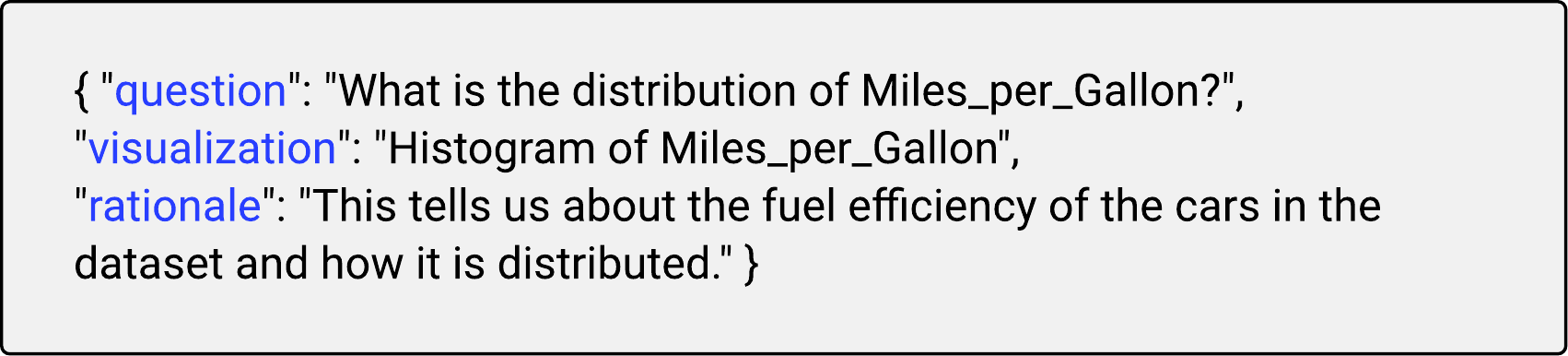}
  \caption{ A goal generated by \lida{} is a JSON data structure that contains a question, a visualization and a rationale. }
  \label{fig:goalex}
\end{figure}

This module generates data exploration goals, given a summary generated by the \summarizer{}. We express goal generation as a multitask generation problem where the \llm{} must generate  a \textit{question} (hypothesis), a \textit{visualization} that addresses the question and \textit{rationale} (see Fig \ref{fig:goalex}).  We find that requiring the \llm{} to produce a rationale leads to more semantically meaningful goals.  

\subsubsection{\visgen{}}

\begin{figure}[htbp]
    \centering
    \includegraphics[width=\columnwidth]{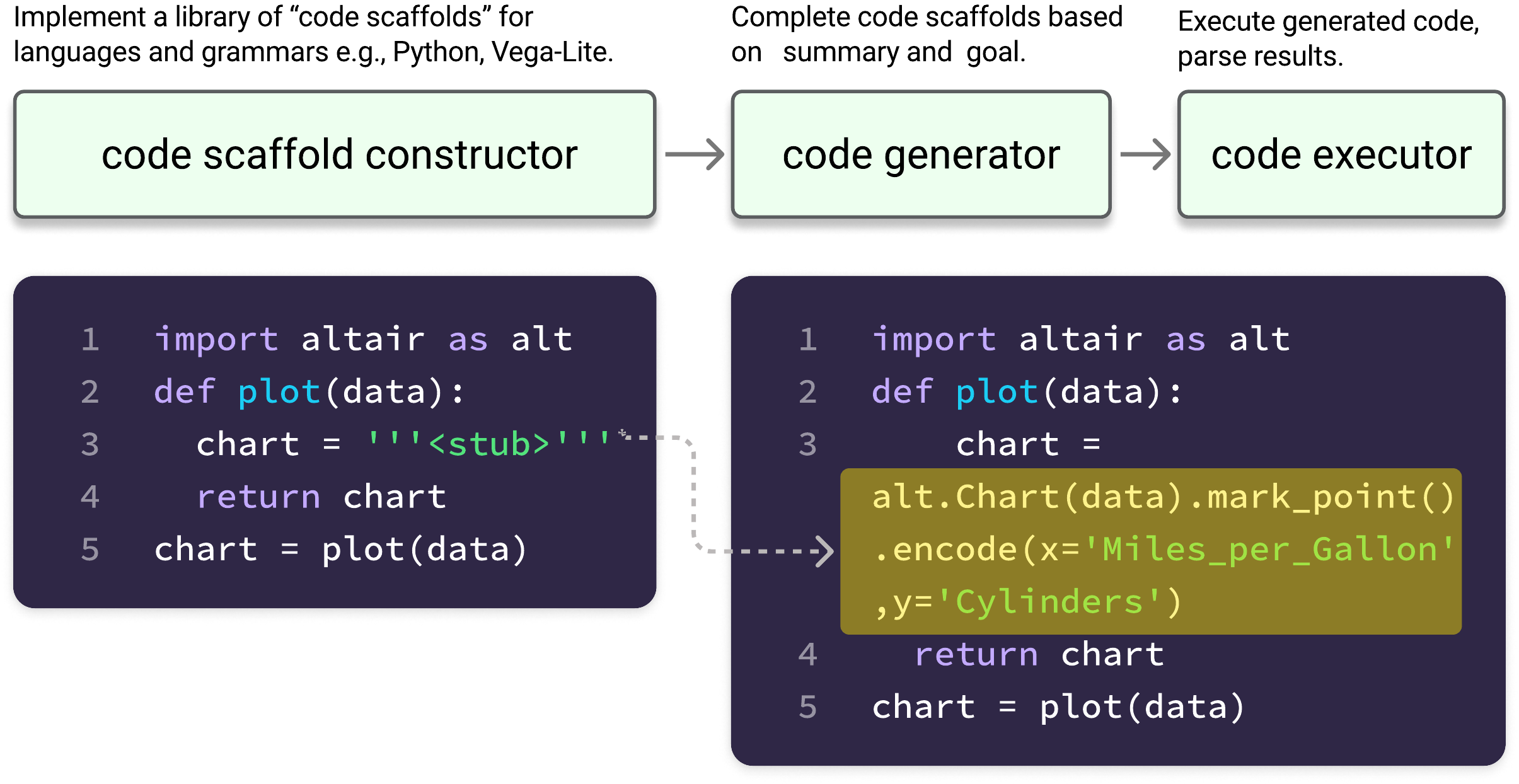}
    \caption{ The \visgen{} module constructs visualization code scaffolds, fills a constrained section ($<stub>$) and executes the scaffold.}
    \label{fig:vizcodegen}
\end{figure}

\noindent The \visgen{} generates visualization specifications and is comprised of 3 submodules - a \textit{code scaffold constructor}, a \textit{code generator} and a \textit{code executor}.

\noindent \textbf{Code scaffold constructor}: Implements a library of code scaffolds that correspond to programming languages and visualization grammars e.g., python scaffolds support grammars such as Matplotlib, GGPlot, Plotly, Altair, Seaborn, and Bokeh. Each scaffold is an \textit{executable program} that i.) imports relevant dependencies ii.) defines an empty function stub which returns a visualization specification (see Fig \ref{fig:vizcodegen}a).

\noindent \textbf{Code generator}: Takes a scaffold, a dataset summary, a visualization goal, and builds a prompt. An \llm{} (applied in \textit{fill-in-the-middle} mode \cite{bavarian2022efficient}) is then used to generate $n$ candidate visualization code specifications.

\noindent \textbf{Code executor}: Post-processes and executes\footnote{Execution in a sandbox environment is recommended.} the code specifications as well as filters the results. \lida{} implements several filtering mechanisms to detect errors, each with latency tradeoffs: 
\begin{enumerate*}[label=(\roman*)]
  \item generates a large sample for $n$ with high temperature, discard candidates that do not compile.
  \item apply self consistency \cite{wang2022self} in \llm{}s where multiple candidates are generated and the solution with the highest consensus is selected.
  \item generate correctness probabilities \cite{kadavath2022language} for all candidates and selects the one with the highest probability.
\end{enumerate*} Note that the last two approaches are computationally expensive (require multiple forward passes through an \llm{}) and are not suitable for real time applications. The final output is a list of visualization specifications (code) and associated raster images.

\subsubsection{ \vizops{} - Operations on Generated Visualizations}
\label{sec:vizops}
Given that \lida{} represents visualizations as code, the \visgen{} also implements submodules to perform operations on this representation.

\noindent \textbf{Natural language based visualization refinement}: Provides a conversational api to iteratively refine generated code (\textit{e.g., translate chart t hindi … zoom in by 50\%} etc) which can then be executed to generate new visualizations.

\noindent \textbf{Visualization explanations and accessibility}: Generates natural language explanations (valuable for debugging and sensemaking) as well as accessibility descriptions (valuable for supporting users with visual impairments).

\noindent \textbf{Visualization code self-evaluation and repair}: Applies an \llm{} to \textit{self-evaluate}  generated code on multiple dimensions (see section \ref{sec:sevq}). 

\noindent \textbf{Visualization recommendation}: Given some context (goals, or an existing visualization), recommend additional visualizations to the user (e.g., for comparison, or to provide additional perspectives).

% \begin{enumerate*}[label=(\roman*)]

%     \item \textbf{Natural language based visualization refinement}. Provides a conversational api to iteratively refine generated code (\textit{e.g., change the x axis to .. translate chart to … zoom in by 50\%} etc) which can then be executed to generate a new visualization.
%     \item \textbf{Visualization explanations and accessibility}. Generates natural language explanations  of the visualization code (valuable for debugging and sensemaking) and accessibility descriptions for the visualizations. 
%     \item \textbf{Visualization code self-evaluation and repair}. Evaluates generated code on multiple dimensions - e.g., code accuracy, transformations, compliance, visualization type, encoding and aesthetics (see section \ref{sec:sevq}).
%     \item  \textbf{Visualization recommendation}. Given some context (goals, or an existing visualization), recommend additional visualizations that may be useful to the user (e.g., for comparison, or to provide additional perspectives or directions).
%   \end{enumerate*} 

\subsection{\infographics{}}
This module is tasked with generating stylized graphics based on output from the \visgen{} module (see Fig \ref{fig:infographics_small}). It implements a library of visual styles described in \nl{} that are applied directly to visualization images. Note that the style library is editable by the user. These styles are applied in generating infographics using the text-conditioned image-to-image generation capabilities of diffusion models \cite{rombach2022high}, implemented using the \textit{Peacasso} library api \cite{dibia2022peacasso}. An optional post processing step is then applied to improve the resulting image (e.g., replace axis with correct values from visualization, removing grid lines, and sharpening edges).

\subsection{\ui{}}
\label{sec:ui}
\lida{} implements a user interface that communicates with the core modules over a REST and Websocket api. The user interface implements several views.   

\noindent \textbf{Data upload and summarization}: This view allows the user to upload a dataset and explore a sample of rows in the dataset via a table view.  A data upload event triggers a call to the \summarizer{} and \goal{} module and displays a summary of the dataset and a list of potential goals. This view also allows the user to optionally annotate and refine the generated summary or curate fields used in the dataset.

\noindent \textbf{Visualization view}: This view allows the user to optionally provide a visualization goal in \nl{} (e.g., "what is the fuel efficiency per country?") or select a generated goal and then displays a generated visualization . For each visualization, intermediate output from the models (underlying data summary, visualization specification, code scaffold) are shown as explanations to aid in sensemaking, and debugging(see Fig \ref{fig:vizgen_view}). This view also implements the \vizops{} capabilities described in Section \ref{sec:vizops} (e.g., See the interface for visualization evaluation in Fig \ref{fig:vizgen_eval}). Note that the \nl{} interface inherits the multilingual language capabilities of the underlying \llm{}, enabling multilingual \nl{} interaction.

Overall, the combination of these modules result in a system that is able to implicitly address an array of data visualization operations such as data transformation, encoding, mark selection, styling, layout, and annotation \cite{wang2022towards}.

\section{Evaluation}
\label{sec:evaluation}
% In this section, we describe the metrics, and experiment settings used for an early evaluation of \lida{}.  

\subsection{Evaluation Metrics}
Our initial evaluation of \lida{} focuses on two high level metrics - visualization error rates (\ver{}) to provide signals on the \textit{reliability} of the \lida{} pipeline, and self-evaluated visualization quality (\sevq{}) to assess the \textit{quality} of generated visualizations.

\subsubsection{Visualization Error Rate (\ver{})}
Visualization error rate is computed as the percentage of generated visualizations that result in code compilation errors. This metric provides critical insights into the reliability of the \lida{} pipeline and impact of changes to the system (e.g., prompt engineering or scaffold update).

\[
\ver{} = \frac{E}{T} * 100
\]

Where:
- \(E\) = Number of generated visualizations with code compilation errors, and 
- \(T\) = Total number of generated visualizations.

\subsubsection{Self-Evaluated Visualization Quality (\sevq{})}
\label{sec:sevq}
Recent work shows \llm{s} like GPT-4 encode broad world knowledge \cite{openai2023gpt4}, can assess the quality of their output \cite{kadavath2022language,lin2022teaching} and can approximate human judgements for tasks such as summarization \cite{liu2023gpteval}. Our observations applying GPT3.5/GPT-4 to visualization tasks suggest similar results. Specifically, GPT-4 has learned to encode \textit{some} visualization best practices and can apply these in generating critiques of visualization code across multiple dimensions. Thus, to evaluate visualization quality, we compute an \sevq{} metric by applying GPT-4 in assessing the quality of generated visualizations. Specifically, we task GPT-4 with scoring generated visualization code (a numeric value from 1-10 and a rationale) across 6 dimensions - code accuracy, data transformation, goal compliance, visualization type, data encoding, and aesthetics. These dimensions are \textit{informed} by existing literature on visualization generation/recommendation e.g., \citet{wang2022towards} outline 6 visualization tasks including data  transformation, encoding, marks, styling, layout and annotation, while \cite{moritz2018formalizing} codify constraints for visualization quality across expressivity (does it convey the facts of the data) and effectiveness (is the information more readily perceived compared to other visualizations) criteria. Additional details on prompts used for each dimension are provided in Appendix \ref{sec:appendix_sevq}.

\subsection{Evaluation Benchmark Settings}
Our initial benchmark is based on 57 datasets sourced from the vega datasets repository\footnote{\url{https://github.com/vega/vega-datasets}}. For each dataset, \lida{} is tasked with generating 5 goals and 1 visualization per goal across multiple grammars\footnote{ \lida{} is given a single try for each step. In theory, the error rates can be driven to zero, by recursively applying the visualization self-evaluation and self-repair modules.}. For reproducibility, we set $temperature=0$ and number of samples $n=1$ for the \llm{}. A gallery of the generated evaluation visualizations can be viewed on the \lida{} \href{https://microsoft.github.io/lida/gallery}{project page}.

\subsection{Evaluation and Ablation Study Results}
\label{sec:eval_results}
\begin{figure}[h]
    \centering
    \includegraphics[width=\columnwidth]{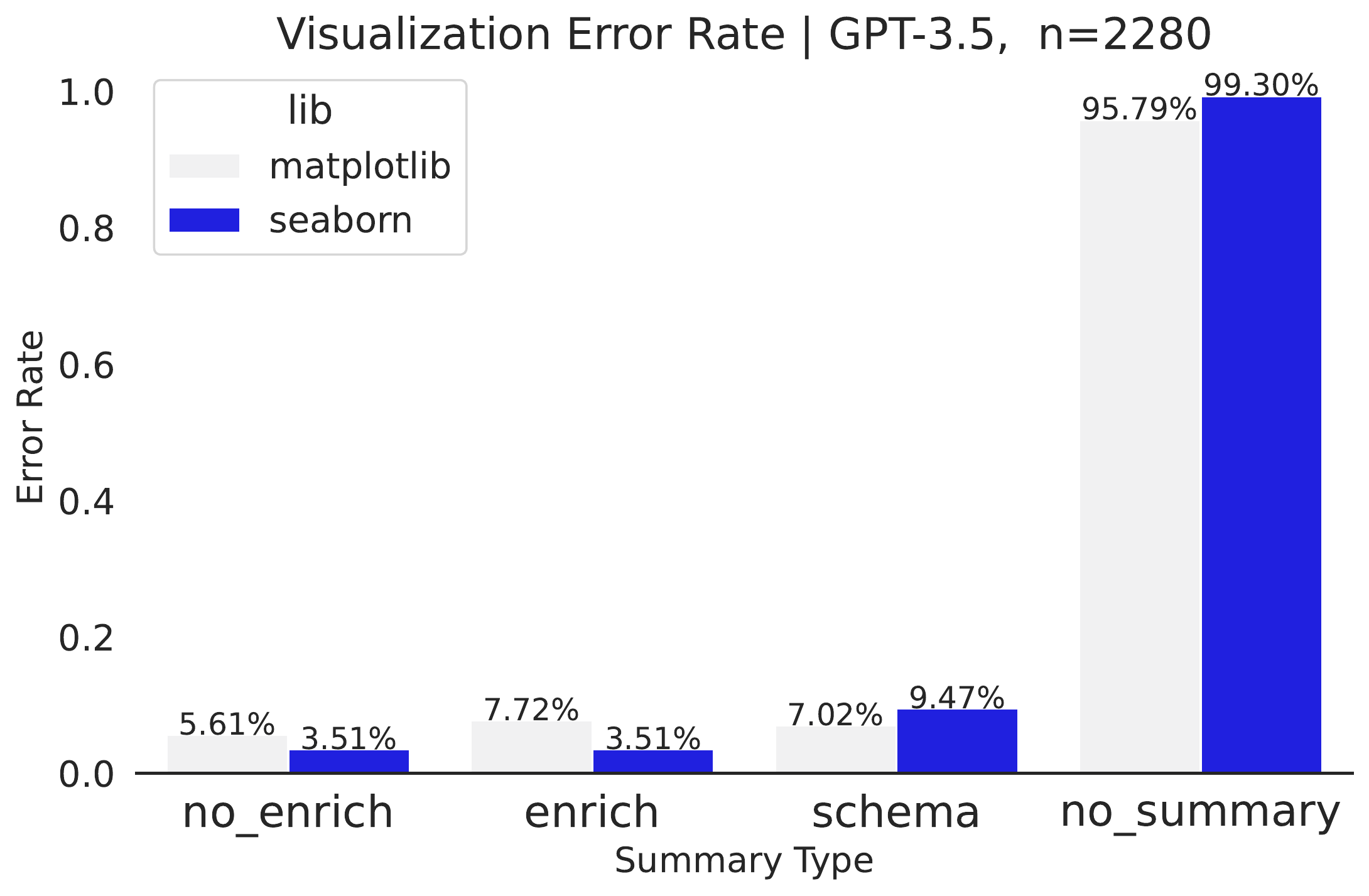}
    \caption{ Results from an ablation study on the impact of data summarization strategies on visualization error rate (\ver{}) metric.
    }
    \label{fig:vizerrorrate}
\end{figure}

Overall, we find that \lida{} is able to generate visualizations with a low error rate (\ver{} = 3.5\%). We also conduct an ablation study to inform on the impact of the \summarizer{} across the following conditions - \begin{enumerate*}[label=(\roman*)]
    \item  \textit{no\_enrich}: a base summary with no enrichment (see Section \ref{sec:summarizer}),
    \item \textit{enrich}: summary with \llm{} enrichment,
    \item \textit{schema}: only field names, i.e., schema as summary, and
    \item \textit{no\_summary}: no summary
  \end{enumerate*}. Results show that including a summary leads to reduced error rate compared to simply adding field names (schema) as summary. We also find that enriching the base summary with an \llm{} has less of an effect on \ver{} (with variations across visualization grammar), and an expressive, well-represented grammar like Seaborn having lower \ver{}. These results are summarized in Figure \ref{fig:vizerrorrate}. We also find that the \sevq{} metric is valuable in identifying semantic quality issues with generated visualizations. For example, Fig \ref{fig:vizgen_eval} shows an example where the user has requested a pie chart, and the \lida{} self-evaluation module critiques this visualization using the \sevq{} metric, providing a rationale for why a bar chart is more effective (see Fig \ref{fig:vizgen_eval}), with the option to automatically repair the visualization.

\section{Conclusion}
\label{sec:conclusion}

In this work, we formulate visualization generation as a multi-stage text (and code) generation problem that can be addressed using large language models. We present \lida{} - a tool for the automatic generation of grammar-agnostic visua\textbf{li}zations an\textbf{d} infogr\textbf{a}phics. \lida{} addresses limitations of current automatic visualization systems - automatic generation of hypothesis/goals given datasets, conversational interface for controllable visualization generation and refinement, support for multiple visualization grammars using the same pipeline and the ability to generate infographics. \lida{} is effective compared to state of the art systems (see example \href{https://microsoft.github.io/lida/gallery}{gallery} of generated visualizations); it offers a simplified system implementation and leverages the immense language modeling and code generation capabilities of \llm{s} in implicitly solving complex visualization subtasks. Finally, we introduce metrics for assessing reliability (visualization error rate - \ver{}) and visualization quality (self-evaluated visualization quality -\sevq{}) for \llm{}-enabled visualization tools. We hope modules implemented in \lida{} will serve as useful building blocks in enabling complex creative workflows such as \textit{visualization translation}, \textit{chart question answering}(with applications in accessibility of charts), automated \textit{data exploration} and \textit{automated storytelling}.

\section{Limitations}

While \lida{} demonstrates clear advances in how we can support users in authoring visualizations and infographics, there are several limitations that offer a natural avenue for future research. 

\noindent \textbf{Low Resource Grammars}: The problem formulation introduced in \lida{} depends on the underlying \llm{}s having \textit{some} knowledge of visualization grammars as represented in \textit{text and cod}e in its training dataset (e.g., examples of Altair, Vega, Vega-Lite, GGPLot, Matplotlib, \textit{represented in}  Github, Stackoverflow, etc.). For visualization grammars not well represented in these datasets (e.g., tools like Tableau, PowerBI, etc., that have graphical user interfaces as opposed to code representations), the performance of \lida{} may be limited without additional model fine-tuning or translation. Furthermore, performance may be limited for complex tasks (e.g., tasks requiring complex data transformations) beyond the expressive capabilities of specific grammars. Further research is needed to: i.) study effects of strategies like task disambiguation  ii.)  impact of task complexity and choice of programing language/grammar  on performance.

\noindent \textbf{Deployment and Latency}: Large language models (e.g., GPT3.5 used in this work) are computationally expensive and require significant compute resources to deploy at low latency. These costs can prove to be impractical for \textit{real-world application}. In addition, the current setup includes a code execution step which is valuable for verification but increases deployment complexity (requires a sandbox). Thus, there is opportunity to: i.) train smaller capable \llm{}s \cite{llamatouvron2023}  finetuned on a curated dataset of programming languages and visualization grammars .ii) design vulnerability mitigation approaches such as limiting program scope or generating only input parameters for visualization grammar compilers.

\noindent \textbf{Explaining System Behavior}: The approach discussed in this paper simplifies the design of visualization authoring systems, but also inherits interpretability challenges associated with large language models. While \lida{} offers intermediate outputs of the model (e.g., generated code and specifications) as \textit{explanations}, as well as post-hoc explanations of generated code (see section \ref{sec:vizops}), there is a need for further research in explaining system behavior (conditions when they are needed) and providing actionable feedback to the user.

\noindent \textbf{System Evaluation}: Benchmarking \llm{}'s on creativity tasks can be challenging. While the current study introduces metrics for evaluating reliability (\ver{}) and visualization quality (\sevq{}) (see section \ref{sec:evaluation}), there is a need for more  comprehensive benchmarks on a variety of datasets and visualization grammars. Furthermore, there are research opportunities to i.) study and quantify the capabilities of \llm{s} in \textit{encoding} and \textit{applying} visualization best practices  ii.) conduct empirical studies that evaluate model behavior, mapping out failure cases and proposing mitigations  iii.) qualitatively study the impact of tools like \lida{} on user \textit{creativity} while authoring visualizations.

\section*{Acknowledgements}
This manuscript has benefited from comments and discussions with members of the HAX group (Saleema Amershi, Adam Fourney, Gagan Bansal), VIDA group (Steven Drucker, Dan Marshall), Bongshing Lee, Rick Barraza and others at Microsoft Research.

\bibliography{paper.bib}
\bibliographystyle{acl_natbib}

\appendix

\section{The \lida{} Library}
\label{sec:appendix_user_interface} 

\begin{figure*}[htbp]
  \centering
  \includegraphics[width=\textwidth]{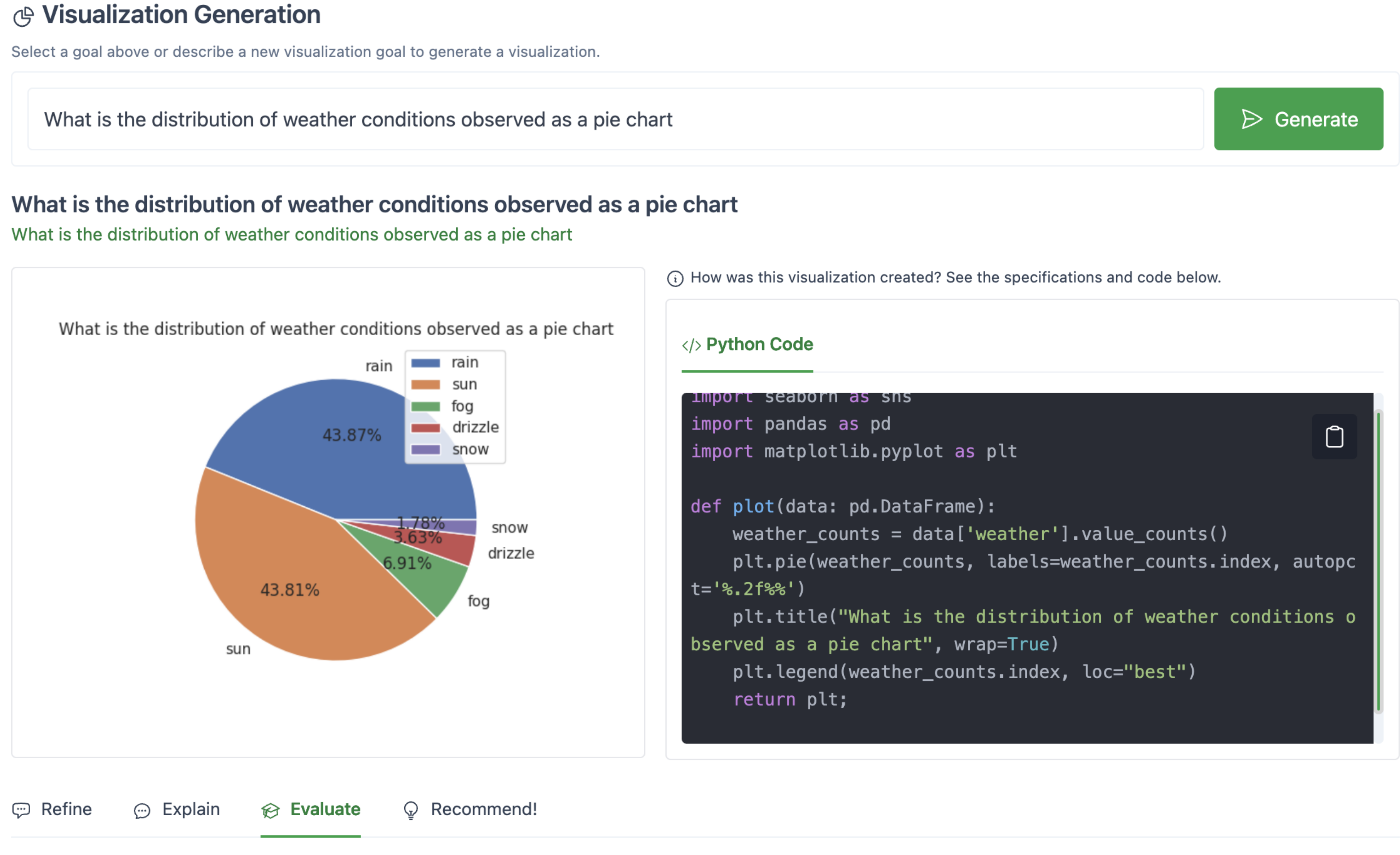}
  \caption{ The visualization generation section of the \lida{} UI enables the user to i.) specify their overall goal in natural language and generate visualizations ii.) inspect, edit and execute generated code iii.) view the generated visualization. iv.) perform operations on generated code e.g., refine, explain, evaluate and recommend visualizations.}
  \label{fig:vizgen_view}
\end{figure*}

\begin{figure*}[htbp]
  \centering
  \includegraphics[width=\textwidth]{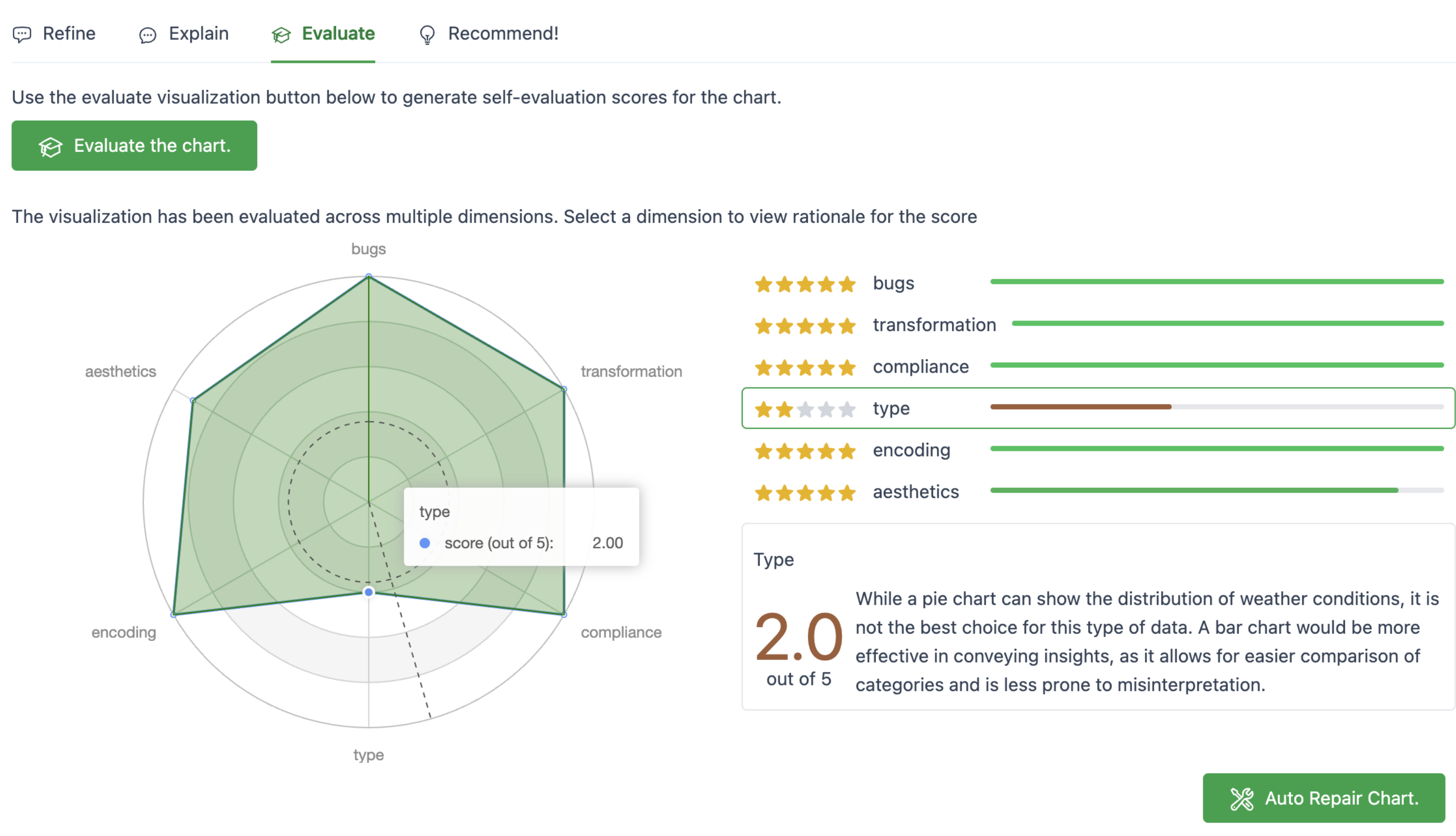}
  \caption{The self-evaluation module in \lida{} is used to evaluate/critique a generated visualization, providing scores across 6 dimensions with rationale. In this case, the visualization contains a pie chart, and a bar chart is recommended as an alternative. }
  \label{fig:vizgen_eval}
\end{figure*}

\lida{} is implemented as a python library with modules for each of the components described in Section  \ref{sec:system}. The library is available on github\footnote{\url{https://github.com/microsoft/lida}} and can be installed using pip - \textit{pip install lida}. The library provides a python api, web api for integration into other applications, and a command line interface. It also provides a web-based user interface for users to interact with \lida{} (Fig \ref{fig:vizgen_eval}, \ref{fig:vizgen_view}). 

\begin{figure}[htbp]
  \centering
  \includegraphics[width=\columnwidth]{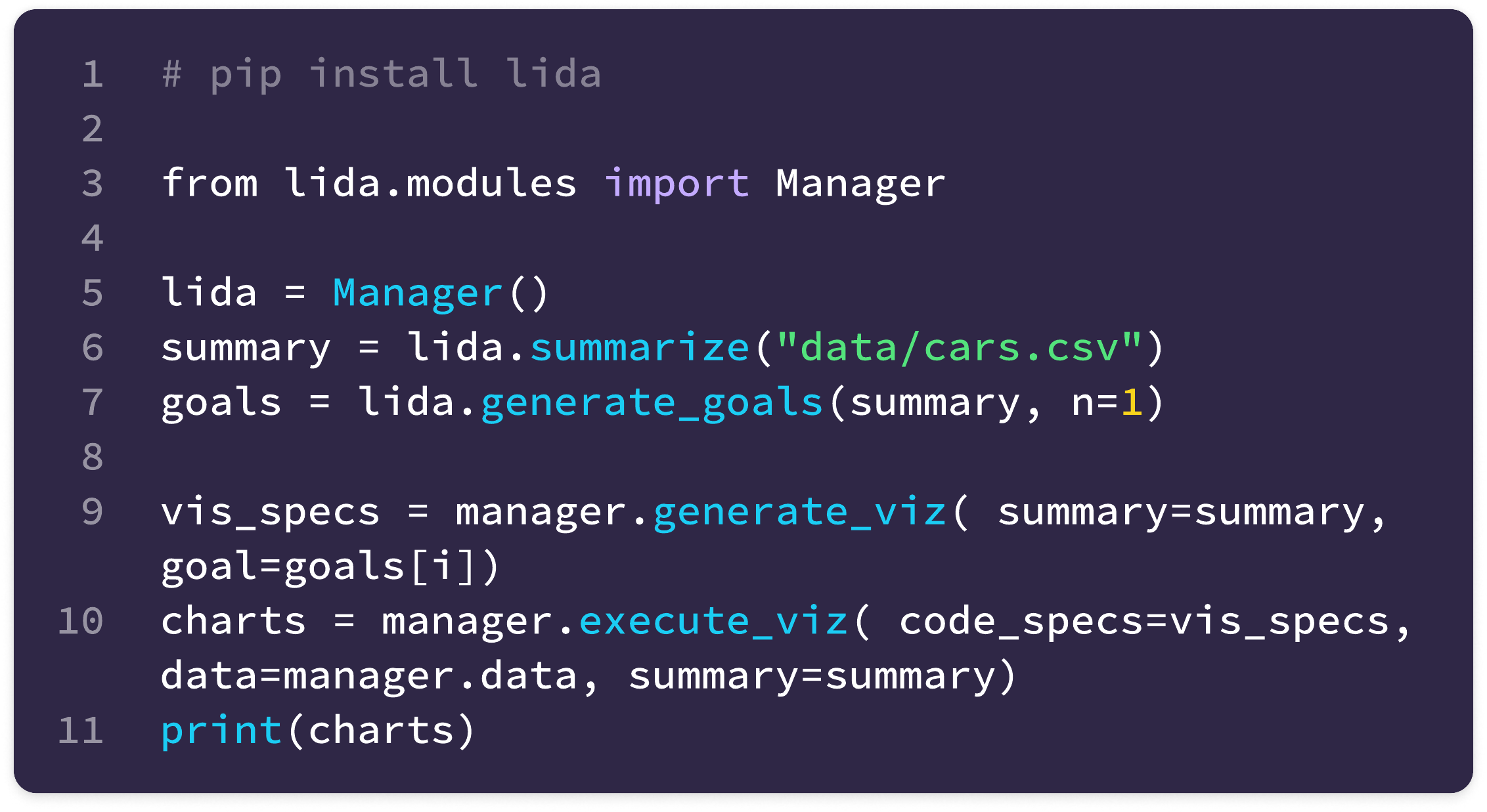}
  \caption{ Example usage of \lida{} shows how to generate a summary, visualization goals,  code specifications and execute the code to generate visualizations.}
  \label{fig:lida_code}
\end{figure}

\begin{figure*}[htbp]
  \centering
  \includegraphics[width=\textwidth]{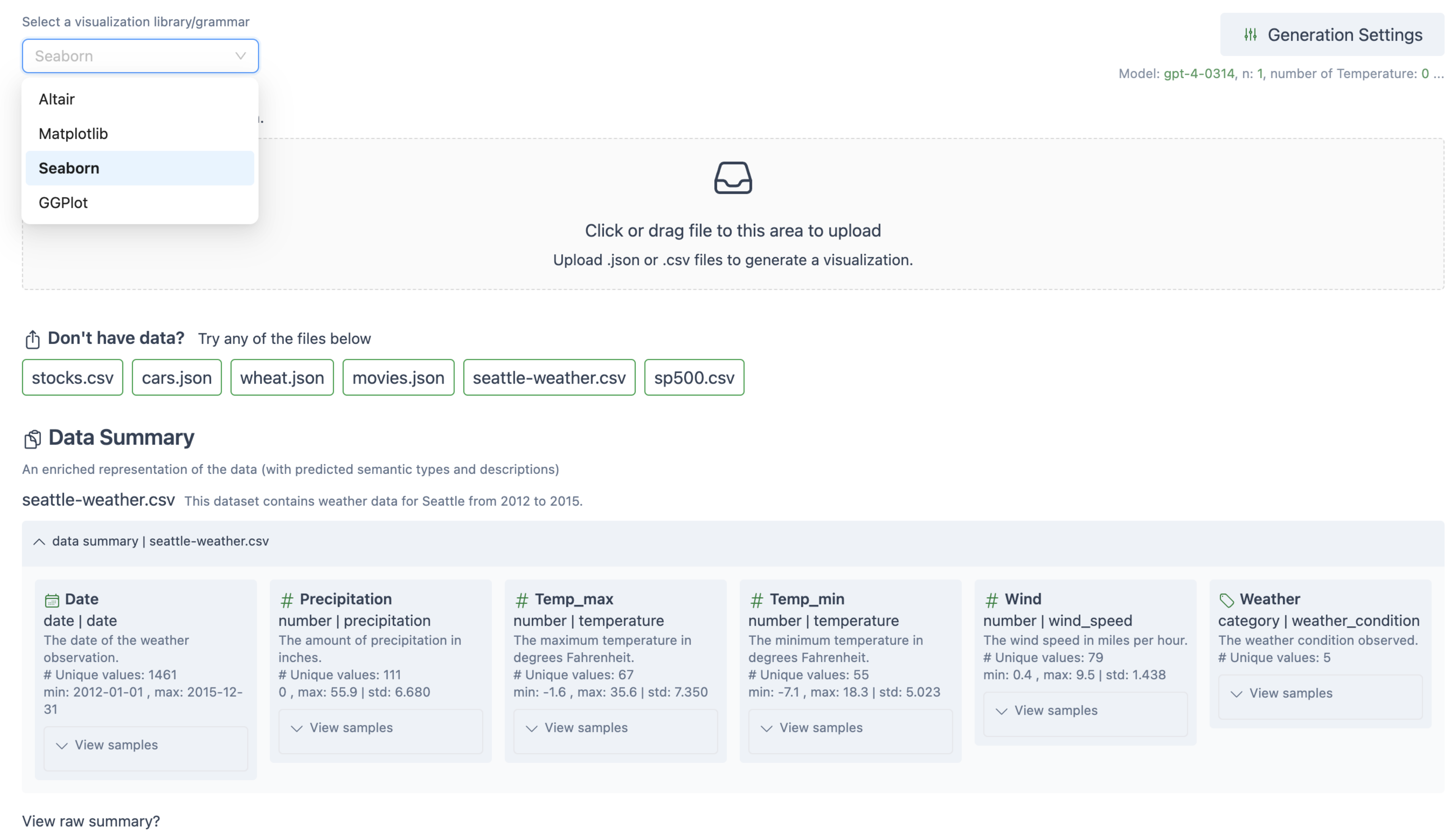}
  \caption{In the data upload section of the \lida{} UI, users can select a grammar of choice and upload a dataset. A dataset upload event triggers a goal generation as well as visualization generation tasks. }
  \label{fig:vizgen_upload}
\end{figure*}

% \begin{figure*}[htbp]
%   \centering
%   \includegraphics[width=\textwidth]{figs/goalexp.pdf}
%   \caption{The goal section of the \lida{} UI shows example goals with question, visualization and rationale. }
%   \label{fig:vizgen_goal}
% \end{figure*}

% \begin{figure*}[h]
%   \centering
%   \includegraphics[width=\textwidth]{figs/vizgenview.png}
%   \caption{ To create a summary of the data, we first generate a rule based summary using the PANDAS library to extract atomic $dtypes$ and general column properties. Next, we apply an LLM to enrich this summary by adding a description of the dataset and columns given the current features.}
%   \label{fig:vizgen_view}
% \end{figure*} 

% \begin{figure*}[h]
%   \centering
%   \includegraphics[width=\textwidth]{figs/vizevalview.png}
%   \caption{ To create a summary of the data, we first generate a rule based summary using the PANDAS library to extract atomic $dtypes$ and general column properties. Next, we apply an LLM to enrich this summary by adding a description of the dataset and columns given the current features.}
%   \label{fig:vizgen_eval}
% \end{figure*}

\section{Self-Evaluated Visualization Quality (\sevq{}) Prompts}
\label{sec:appendix_sevq}

For the \sevq{} metric, we use GPT-4 to assess visualization quality by scoring generated visualization code across the 6 task dimensions - code accuracy, data transformation, goal compliance, visualization type, data encoding, and aesthetics. These dimensions are implemented as prompts to an LLM \footnote{Exact prompts can be found at the project repository \url{https://github.com/microsoft/lida}.}, which then generates a score between 1-10 for each dimension. The final \sevq{} score is the average of the 6 scores. A sketch of the prompts used for each dimension are enumerated in table \ref{table:evaluation-dimensions}.

% \noindent \textbf{code accuracy}: \textit{``Does the code contain bugs, logic errors, syntax error or typos? How serious are the bugs? How should it be fixed?''} \\

% \noindent \textbf{data transformation}: \textit{``Is the data transformed appropriately for the visualization type?''} \\

% \noindent \textbf{goal compliance}: \textit{``How well the code meets the specified visualization goals?''} \\

% \noindent \textbf{visualization type}: \textit{``Considering best practices, is the visualization type appropriate for the data and intent? Is there a visualization type that would be more effective in conveying insights?''} \\

% \noindent \textbf{data encoding}: \textit{``Is the data encoded appropriately for the visualization type?''} \\

% \noindent \textbf{aesthetics}: \textit{``Are the aesthetics of the visualization appropriate and effective for the visualization type and the data?''} \\

\begin{table}[h]
  \centering
  \small % added small command for reduced font size
  \begin{tabular}{p{0.2\linewidth}p{0.7\linewidth}}
  \toprule
  \textbf{Dimension} & \textbf{Prompt} \\
  \midrule
  Code accuracy       & Does the code contain bugs, logic errors, syntax error or typos? How serious are the bugs? How should it be fixed? \\
  \midrule
  Data transformation & Is the data transformed appropriately for the visualization type? \\
  \midrule
  Goal compliance     & How well the code meets the specified visualization goals? \\
  \midrule
  Visualization type  & Considering best practices, is the visualization type appropriate for the data and intent? Is there a visualization type that would be more effective in conveying insights? \\
  \midrule
  Data encoding       & Is the data encoded appropriately for the visualization type? \\
  \midrule
  Aesthetics          & Are the aesthetics of the visualization appropriate and effective for the visualization type and the data? \\
  \bottomrule
  \end{tabular}
  \caption{Summary of the evaluation dimensions and the corresponding prompt sketches.}
  \label{table:evaluation-dimensions}
  \end{table}

\section{Design Reflections}
\label{sec:reflection}
Building a system that leverages foundation models (text and images) involves engineering decisions across a wide design space. In this section, we briefly reflect on some of the design choices we made for \lida{} components and the tradeoffs we considered.  

\subsection{Prompt Engineering}
We explored multiple approaches to building prompts that maximized the probability of the \llm{} solving each subtask.

% \subsubsection{Prompt Design}: 

\begin{itemize}
        \item \summarizer{}: We found that improving the  richness of the summary (qualitative \nl{} description, including semantic types) was \textbf{\textit{critical}} to improved quality of generated goals and visualization code. Implementation wise, we began with a manually crafted summary of the data (see Section \ref{sec:summarizer}), and then enriched it via calls to an LLM \textit{and} optional user refinement of the summary.
        \item \goal{}: Providing few shot examples in the prompts where fields and rationale are linked via symbols (e.g., plot a histogram of field X vs Y to show relationship between X and Y) nudges the model to use exact dataset field names, and minimizes the occurrence of hallucinated fields. Prompt engineering also provides mechanisms to bake in visualization best practices e.g. \textit{avoid pie charts}, \textit{apply visualization best practices}, \textit{Imagine you are a highly experienced visualization specialist and data analyst}.
        \item \visgen{}: Casting visualization code generation as a \textit{fill-in-the-middle} problem (as opposed to free-from completion) ensures the model to generates executable code \textit{focused} on the task. For example, in Fig \ref{fig:vizcodegen}, the model is \textit{instructed} to generate only the $<stub>$ portion of the code scaffold. We also note that the degrees of freedom alloted to the model (e.g., specifying how much of the scaffold to complete) can influence its ability to add tasks with varied complexity. For example, a scaffold that allows the model generate data preprocessing code (and includes libraries like statsmodels etc) allows the model to address tasks that require steps such as data transformation, sampling and statistical analysis before generating visualizations etc. 
        \item Overall, we found that setting a low temperature ($t=0$;  generating the most likely visualization) coupled with a per-grammar code scaffold provided the best results in terms of yielding code that correctly compiles into visualization specifications and faithfully addresses the subtask. We also explored prompt formulations that addressed multiple tasks to minimize costs (latency and compute). For example, summary enrichment is a single call where the \llm{}  must generate dataset descriptions, field descriptions and semantic types.
      \end{itemize}

\subsection{Infographic Generation}

We found that setting a low $strength$ parameter ($0.25 < \textit{strength} < 0.45$) for the latent diffusion model (image-to-image mode) and using parsimonious style prompts resulted in stylized images that were faithful to the general \textit{structure} of the original visualization, minimizing distorted or irrelevant imagery. This sort of controlled generation is \textit{necessary} to avoid the distraction \cite{haroz2015isotype} that can arise from superfluous imagery in infographics.

\subsection{Natural Language Interaction}
\begin{enumerate*}[label=(\roman*)]
    \item  \textsc{Hybrid Interface}: Providing a hybrid interface that allows traditional direct manipulation steps in creating visualizations (e.g., selecting which fields to use), paired with a \nl{} interface allows users to leverage existing mental models with traditional visualization tools as well as the \nl{} affordances of \lida{}.
    \item \textsc{\nl{} Interaction Modes}: Beyond generating a base visualization, we also enable operations on generated visualization code (e.g., refinement, explanation, evaluation, recommendation). This builds on insights from \citet{mitra2022facilitating} who propose multi-turn dialog interfaces for visualization authoring towards resolving ambiguities.
  \end{enumerate*}

\begin{figure*}[htbp]
  \centering
  \includegraphics[width=\textwidth]{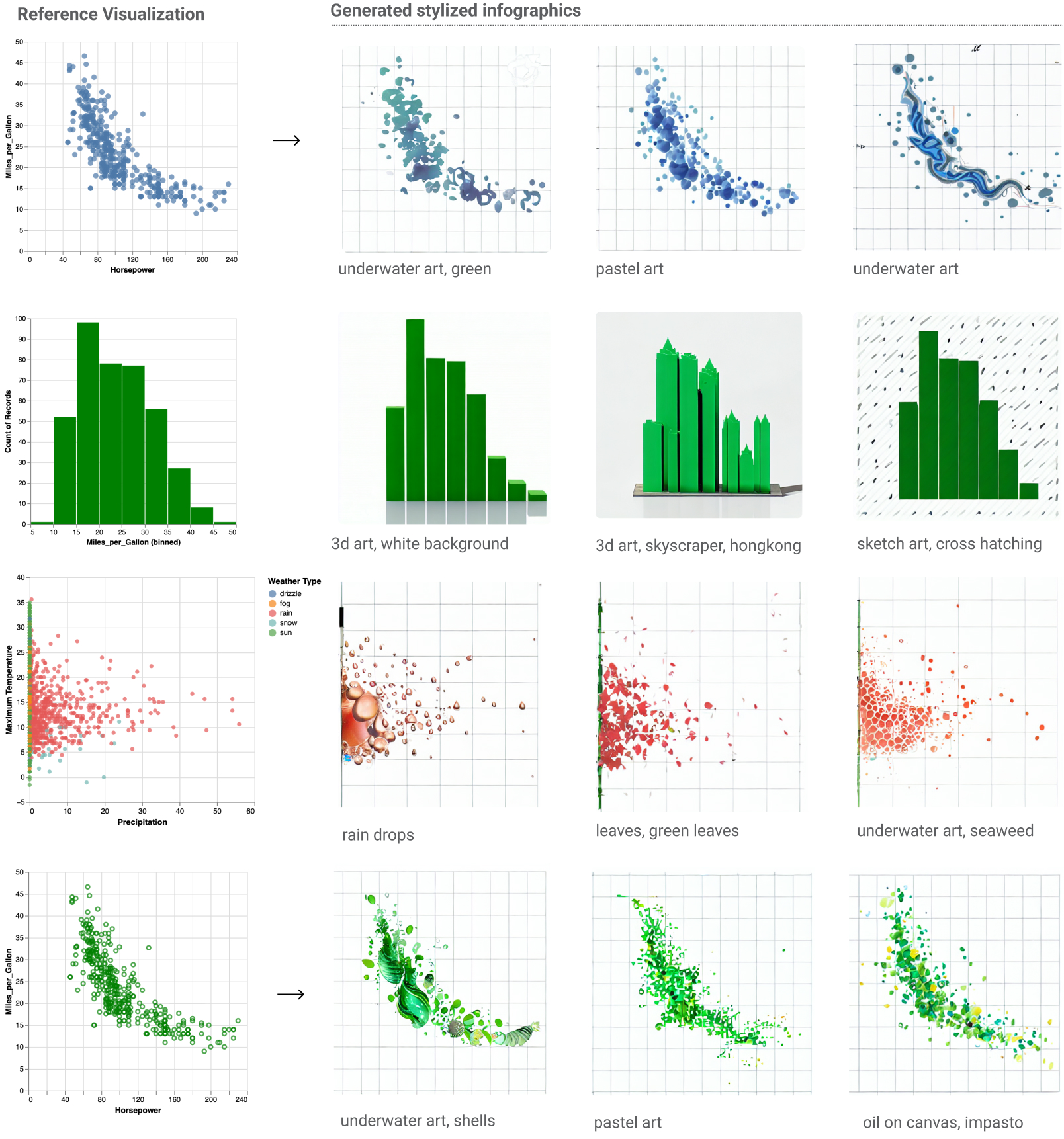} 
  \caption{ The \lida{} infographer module supports the generation of data-faithful infographics. Each infographic is conditioned on a generated visualization as well as natural language style tags which can be used to customize the appearance of the chart.}
  \label{fig:infrographicsmore}
\end{figure*}

\end{document}